%% file: main.tex
\documentclass[journal]{IEEEtran}
\IEEEoverridecommandlockouts
\usepackage{graphicx}
\usepackage{etoolbox}
\usepackage{nicematrix}
\usepackage{tikz}
\usepackage{xspace}
\usepackage{bm}
\usepackage{unicode}
\usepackage[flushleft]{threeparttable}
\usepackage{makecell}
\usepackage{enumitem}
\usepackage{mwe}
\usepackage[fontsize=10.5pt]{fontsize}
\usepackage{hyperref}
\usepackage{academicons}
\usepackage[column=0]{cellspace}
\usepackage{subcaption}
\usepackage{multirow}

\begin{document}

\title{Precision in Building Extraction: Comparing Shallow and Deep Models using LiDAR Data.}

\author{Muhammad Sulaiman$^1$*\href{https://orcid.org/0000-0003-3962-2635}{\includegraphics[scale=0.01]{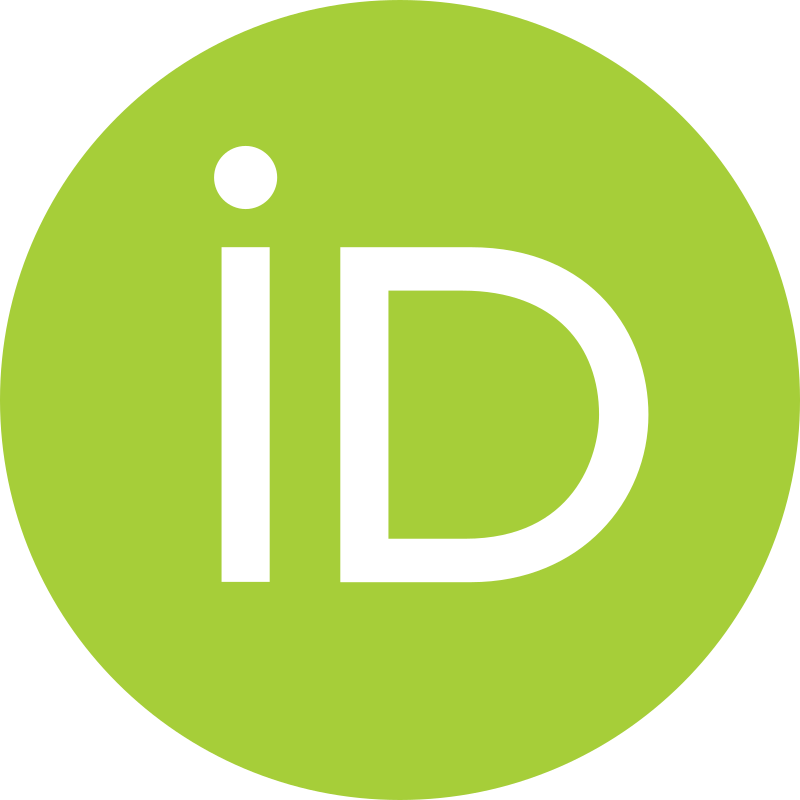}}, Mina Farmanbar$^{1}$, Ahmed Nabil Belbachir$^2$ Chunming Rong$^1$\\
$^1$Department of Electrical Engineering and Computer Science, University of Stavanger, Norway \\
$^2$NORCE Norwegian Research Centre, Norway\\
*Corresponding author: muhammad.sulaiman@uis.no}

\maketitle
\begin{abstract}
Building segmentation is essential in infrastructure development, population management, and geological observations. This article targets shallow models due to their interpretable nature to assess the presence of LiDAR data for supervised segmentation.  The benchmark data used in this article are published in NORA MapAI competition for deep learning model. Shallow models are compared with deep learning models based on Intersection over Union (IoU) and Boundary Intersection over Union (BIoU). In proposed work, boundary masks from the original mask are generated to improve the BIoU score, which relates to building shapes' borderline. The influence of LiDAR data is tested by training the model with only aerial images in task 1 and a combination of aerial and LiDAR data in task 2 and then compared. shallow models outperform deep learning models in IoU by 8\% using aerial images (task 1) only and 2\% in combined aerial images and LiDAR data (task 2). In contrast, deep learning models show better performance on BIoU scores. Boundary masks improve BIoU scores by 4\% in both tasks. Light Gradient-Boosting Machine (LightGBM) performs better than RF and Extreme Gradient Boosting (XGBoost). 

\end{abstract}

\begin{IEEEkeywords}
Building Extraction, Machine Learning, LiDAR data.
\end{IEEEkeywords}

\section{Introduction}
\input{chapters/chapter1}

\section{Literature}
\input{chapters/chapter2}

\section{Methodology}
\label{relatedwork}
\input{chapters/chapter3}

\section{Experiments}
\input{chapters/chapter4}

\section{Discussion}
\input{chapters/chapter5}

\section*{Acknowledgment}
This work is supported by the project EMB3DCAM “Next Generation 3D Machine Vision with Embedded Visual Computing” and co-funded under the grant number 325748 of the Research Council of Norway.

\bibliographystyle{IEEEtran}
\bibliography{main}

\end{document}

%% file: chapters/chapter1.tex
\vspace{0.5em}
Buildings play an essential role in planning policies related to infrastructures and provide data related to population, which helps in management \cite{grecea2013cadastral}. geological observation technologies, such as satellites and drones provide high spatial resolution images and are used in building inspection \cite{fu2019dual}. Easy access to data repository makes it possible to work on different applications, like population aggregation ~\cite{xu2018automatic,xie2015population,cao2020multi}, urban planning \cite{rathore2016urban}, building model reconstruction \cite{rottensteiner2003automatic}, mapping \cite{liu2019building}, emergency rescue \cite{blaschke2010object}, and pre-disaster building risk assessment \cite{hu2021automated}. Manual interpretation and vectorization were difficult and time-consuming and impossible for a big dataset of images. The rapid development of sensors such as Light Detection and Ranging (LiDAR) \cite{cao2020multi}, Polarimetric Synthetic Aperture Radar (POLSAR) \cite{deng2019improved}, and Synthetic aperture radar (SAR) \cite{huang2019automatic} provides enriched data for the automatic extraction of buildings. Computer vision provides different methods like object detection and segmentation for automation in several applications like urban planning and disaster recovery \cite{sun2020applications}. Apart from data availability and image processing techniques, the high spectral and textural similarity between buildings, background objects and shadows of the buildings, and various shapes and colors in the building make this automation challenging.

Automatic building extraction from remote sensing information is essential to determine pre-disaster management in rural and urban areas. Researcher have tried different traditional and deep learning algorithms to improve building extraction ~\cite{schlosser2020building,huang2020automatic,liu2019automatic,dong2018extraction}. The buiding extraction methods rely on features such as building color \cite{cote2012automatic}, specturm \cite{zha2003use}, texture \cite{zhang2017urban}, shape \cite{sun2015regular}  and context ~\cite{huang2011multidirectional,kanwal2023detection}. However buildings having diverse roof colors and textures, lighting, and shadow problem due to weather, still need to work on creating a stable model for generalized results \cite{shao2020brrnet}. LiDAR is independent of spectral and spatial information i.e., shadow, dark or lightening \cite{zhang2006automatic}, and the depth information provided by LiDAR is quite handy to extract ground objects ~\cite{huang2019automatic,niemeyer2014contextual} and improve building extraction results on remote sensing images. Furthermore combining the visual information of optical images and the depth information of LiDAR data can further improve the building extraction task as compared to individual optical images or LiDAR data. Fusion of optical image and LiDAR data requires sensor alignment, and data acquisition is usually costly as compared to single source data.

Pixel-based and object-oriented image classification are two common methods for building extractions \cite{huang2019automatic}. Pixel-based methods can improve performance by combining spectral features and point cloud information \cite{khoshelham2010performance}. Pixel-based image segmentation could be done using both conventional machine learning and deep learning methods. This article focused on conventional machine learning methods to extract buildings and find the significance of LiDAR data. 

\subsection{Dataset}

\begin{figure*}
     \centering
     \begin{subfigure}[b]{0.328\textwidth}
         \centering
         \includegraphics[width=\textwidth]{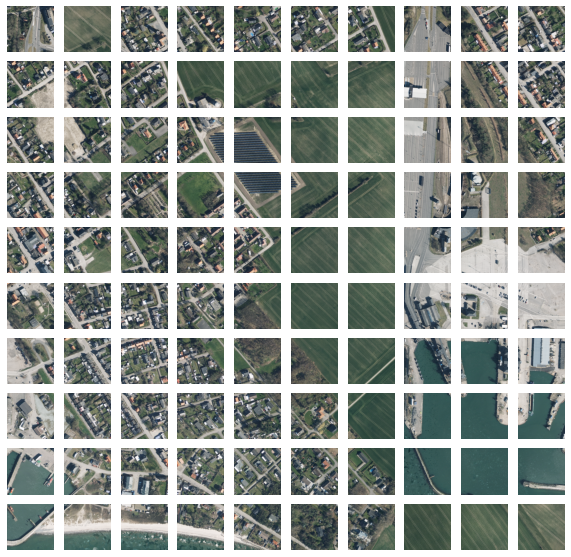}
         \caption{Aerial image}
     \end{subfigure}
     \hfill
     \begin{subfigure}[b]{0.328\textwidth}
         \centering
         \includegraphics[width=\textwidth]{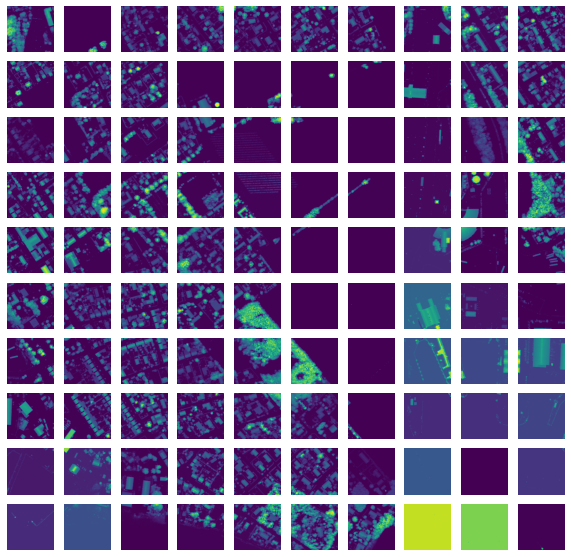}
         \caption{LiDAR data}
     \end{subfigure}
     \hfill
     \begin{subfigure}[b]{0.328\textwidth}
         \centering
         \includegraphics[width=\textwidth]{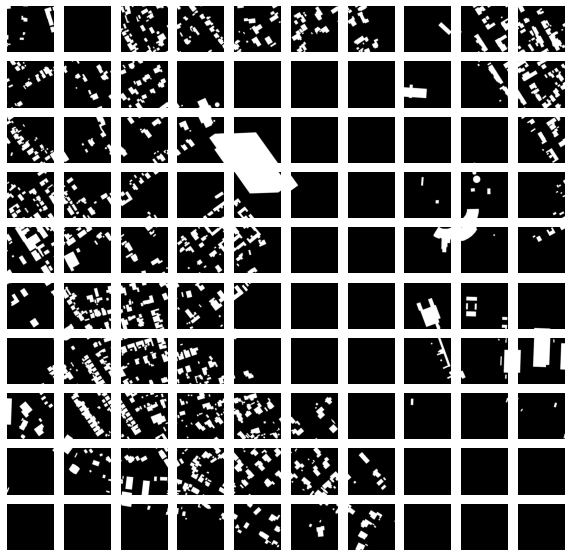}
         \caption{Mask}
     \end{subfigure}
        \caption{Image resolution 5000x5000 pixels is divided into 100 images with equal resolution 500x500 pixels.}
        \label{dataset}
\end{figure*}

This work uses a dataset from NORA's competition named "MapAI: Precision in building segmentation" \cite{jyhne2022mapai}. The dataset consists of real-world data having noise in different forms, different quality images, and large class imbalances. The dataset consists of training, evaluation, and testing images from different regions in the form of aerial images, LiDAR data, and masks. A single image's resolution in aerial and LiDAR is 500X500 pixels. LiDAR data is preprocessed and converted to a matrix like an image, where each value represents the pixel's depth from the LiDAR. The training dataset consists of images from different locations in Denmark, and the test dataset consists of seven locations in Norway. The competition is based on two tasks: 1) Classify buildings from the ground in aerial images and 2) use aerial and LiDAR data. In the second task, the fusion of aerial images with LiDAR data is allowed. Figure \ref{dataset} show first 100 images from the training set. NORA's competition score is divided into two task. Task 1 is "Aerial Image Segmentation Task" where only aerial images are allowed for training and Task 2 is "Laser Data Segmentation Task" where the model could be trained using LiDAR data with or without aerial images \cite{jyhne2022mapai}.

\subsection{Evaluation Metrics}

Image segmentation could be evaluated using region-based and boundary-based metrics. Intersection over Union (IOU) or Jaccard Index (JI) could be used as region-based metrics, which measures the similarity between two binary images, one is ground truth $I_g$, and the other is predicted mask  $I_p$ by dividing the intersection area by total area shown in Equation \ref{eq:iou}  \cite{cho2021weighted}. Boundary Intersection over Union (BIOU) is used as a metric for boundary-based evaluation. BIOU is the intersection over union of the edged ground truth and edged prediction mask, where d denotes the thickness of the edge from the contour line in Equation \ref{eq:biou} \cite{cheng2021boundary}.
\begin{equation}
\label{eq:iou}
IoU = JI = \frac{Intersection}{Union}  = \frac{|I_g \cap I_p|}{|I_g|+|I_p|-|I_g \cap I_p|}  
\end{equation}
\begin{equation}
\label{eq:biou}
BIoU = \frac{|(I_{g_d} \cap I_g)\cap(I_{p_d} \cap I_p)|}{|(I_{g_d} \cap I_g)\cup(I_{p_d} \cap I_p)|}
\end{equation}

Dataset is trained and tested on shallow models and compared with deep learning to show the difference in performance.
Different filters are interpreted using RF and XGBoost to find the best filters for the given dataset. Boundary masks are created to improve BIoU, and each model is trained with and without a boundary mask to compare the differences. Models are tested on data with and without LiDAR data to find the influence of LiDAR data.

%% file: chapters/chapter2.tex

\vspace{0.5em}
The Random Forest (RF) algorithm, was first introduced by \cite{breiman2001random} and has now grown to a standard non-parametric classification and regression tool for constructing prediction rules based on various types of predictor variables without making any prior assumption on the form of their association with the response variable. Neural networks, the basis of Deep Learning (DL) algorithms, have been used in the remote sensing community for many years. Deep learning methods have the ability to retrieve complex patterns and informative features from satellite image data. However, before the development of DL, the remote-sensing community had shifted its focus from neural networks to Support Vector Machine (SVM) and ensemble classifiers, e.g., RF, for image classification and other tasks (e.g., change detection) \cite{ma2019deep}. Results from \cite{maimaitijiang2020soybean} agree with the previous studies ~\cite{zhang2018mu,cai2018high}, which demonstrated that DNN was only slightly superior to SVM and RF in classification and estimation applications.

However, one of the main problems with deep learning approaches is their hidden layers and “black box” nature \cite{habib2009support}, which results in the loss of interpretability. Due to the black-box nature of deep learning, it is impossible to measure the significance of LiDAR data. In contrast, RF and XGBoost are interpretable in nature, and easy to assess the importance of LiDAR data for segmentation. Another limitation of deep learning models is that they are highly dependent on the availability of abundant high-quality ground truth data. On the other hand, recent research works show SVM and RF (i.e., relatively easily implementable methods) can handle learning tasks with a small amount of training dataset yet demonstrate competitive results with Convolutional Neural Networks (CNNs) \cite{mboga2017detection}. Although there is an ongoing shift in the application of deep learning in remote sensing image classification, SVM and RF have still held the researchers’ attention due to lower computational complexity and higher interpretability capabilities compared to deep learning models. RF in terms of classification accuracies, makes it the most popular machine learning classifier within shallow models for remote sensing community \cite{mather2016classification}.

%% file: chapters/chapter3.tex
\vspace{0.5em}
In this work, RF, XGBoost, and LightGBM are used as pixel classifiers to employ segmentation and compare performance with deep learning models tested on the same dataset. RF is a collection of a bunch of decision trees. Decision trees are well known for interpretability and representability, as they mimic how the human brain makes decisions. Interpretability may reduce prediction accuracy. However, ensembles of decision trees overcome this problem and proposed a strong and robust model in the form of RF, and later some extensions in the form of XGBoost and LightGBM.

Bagging train multiple trees on different subsets of datasets having all features and then predict out the label using the average or majority vote of these trees. As an extension of bagging, along with a random subset of the dataset, RF used a random selection of features for each tree, which helps interpretability. However, in RF, trees are independent which avoid the usage of knowledge from the previous learner or tree.

Boosting overcome this independency problem of trees in bagging and building an ensemble of consecutive trees where each tree uses a residual from the previous tree to minmize loss \cite{khoshgoftaar2010comparing}. An extension of Boosting, Gradient Boosting use both gradient descent and boosting to optimize loss function. An extreme version of Gradient Boosting is XGBoost, which is more efficient, flexible, and portable due to advanced regularization which improve generalization.

LightGBM is another version of Gradient Boosting which more focus on computational efficiency and performance as compared to XGBoost. Light GBM reduces the number of splits by employing leaf-wise split rather than level-wise split \cite{ke2017lightgbm}. The remaining part of the section divides into two subsections: Feature Extraction and Segmentation.

\subsection{Feature Extraction}

Preprocessing steps are employed in this work to prepare data for training the model. The dataset consists of both aerial images and LiDAR data. In the first step, blue, green, and red channels are extracted from the aerial image along with the gray image as features of the image. LiDAR data in the dataset is a 2D array having a dimension (500X500) the same as the aerial image. LiDAR data is also fused with other features due to the same dimension, to exploit the presence of LiDAR data for segmentation. 

\begin{figure}
\centering
\includegraphics[width=\columnwidth]{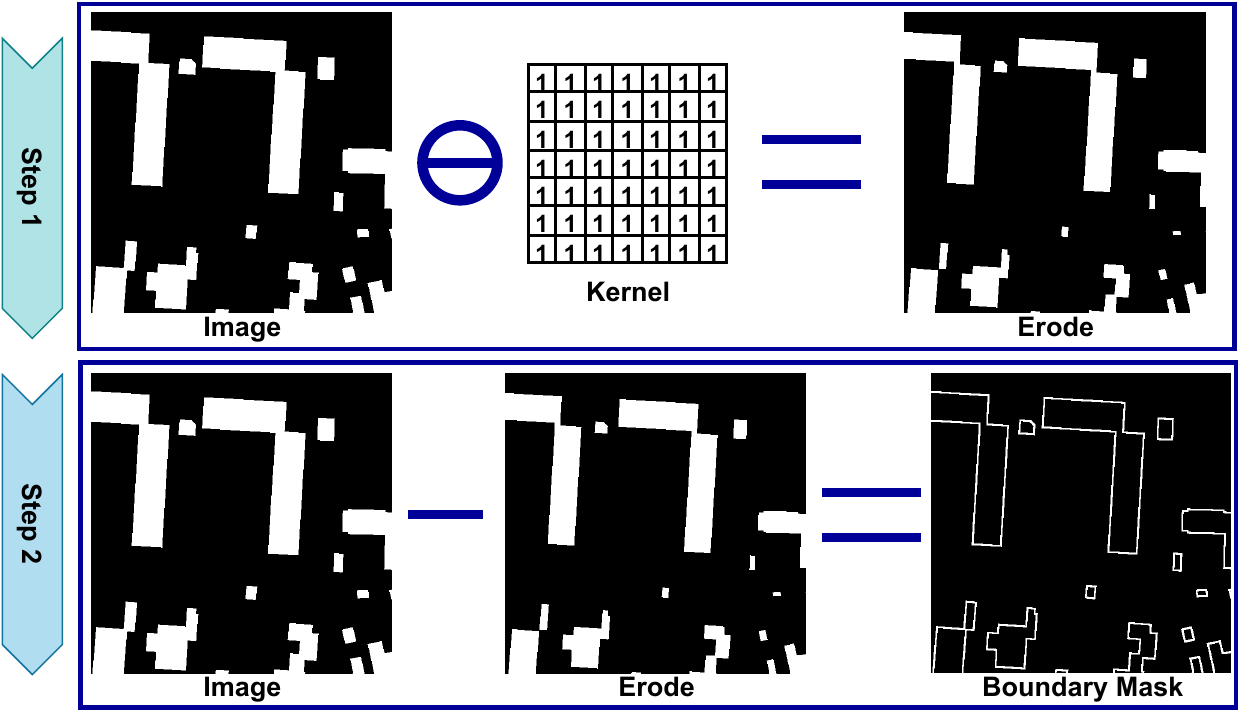}
\caption{BIOU Mask generation: First step Eroding, and Second step Subtracting eroded mage from original image} \label{BIOUMask}
\end{figure}

In the second step, boundary masks are created to improve the BIOU metrics. 7X7 kernel is used as a structuring element to erode the original image, which reduces 3 pixels from each side of all shapes in the image. Figure \ref{BIOUMask} shows the procedure for creating a BIOU mask. In the first step, the image is eroded with a structuring element filled with 1's to erode shapes in the image equally from all sides, which results in an eroded image. In the second step eroded image is subtracted from the original image, which results in a BIOU mask.

\begin{figure*}
\includegraphics[width=\textwidth]{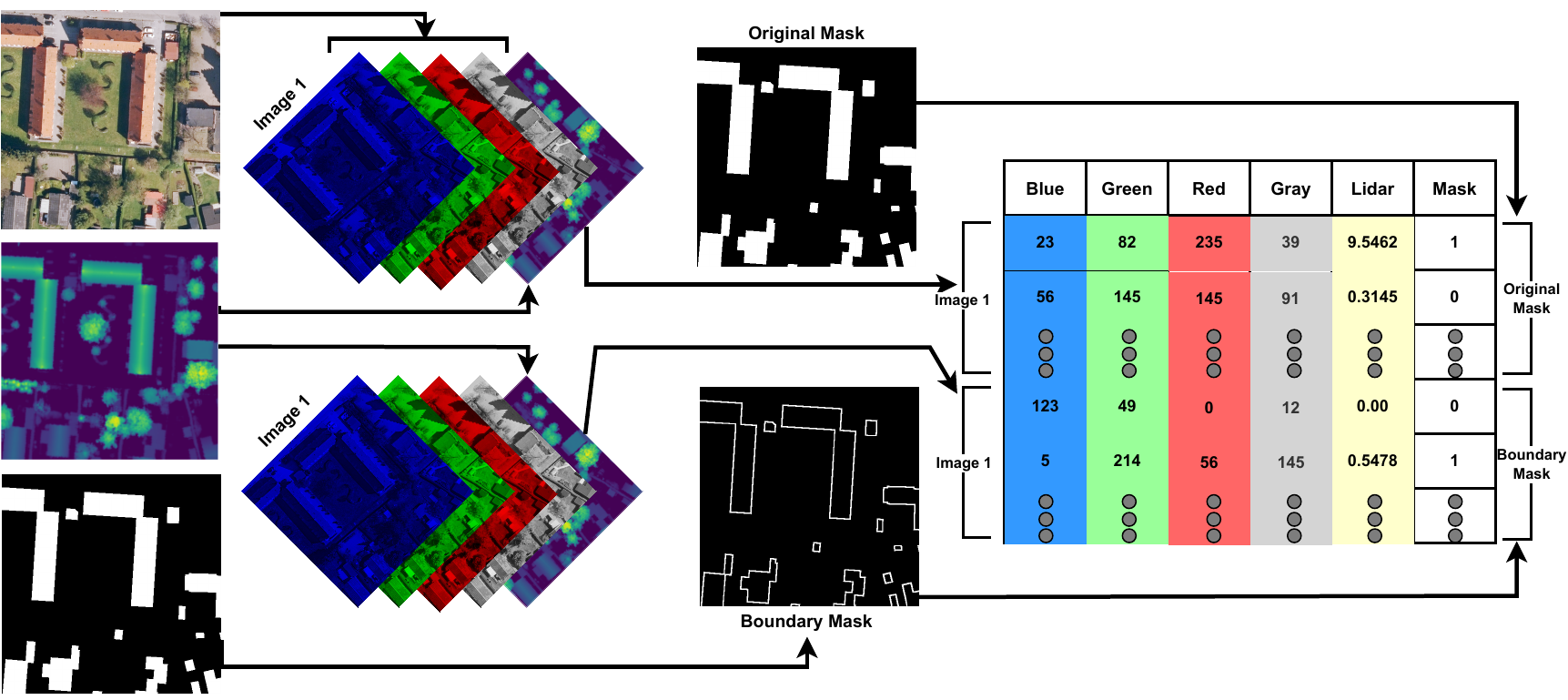}
\caption{Feature Extraction and Data Preprocessing} \label{Preprocessing}
\end{figure*}

In the third step, features are aligned to train the model. Blue, green, red, gray, and LiDAR features of the first image having 500X500 dimensions are Flattered and placed in the matrix with the original mask as a label. Hence the first tuple of the matrix consists of the first pixel values for blue, green, red, gray, LiDAR, and mask as a label. Same features of the first image are duplicated for the boundary mask. In this way, the model is trained with both the original mask and boundary mask to improve the BIOU metric along with IOU. Figure \ref{Preprocessing} shows the data preprocessing procedure for this work.

\subsection{Segmentation}
\begin{figure}
\centering
\includegraphics[width=\columnwidth]{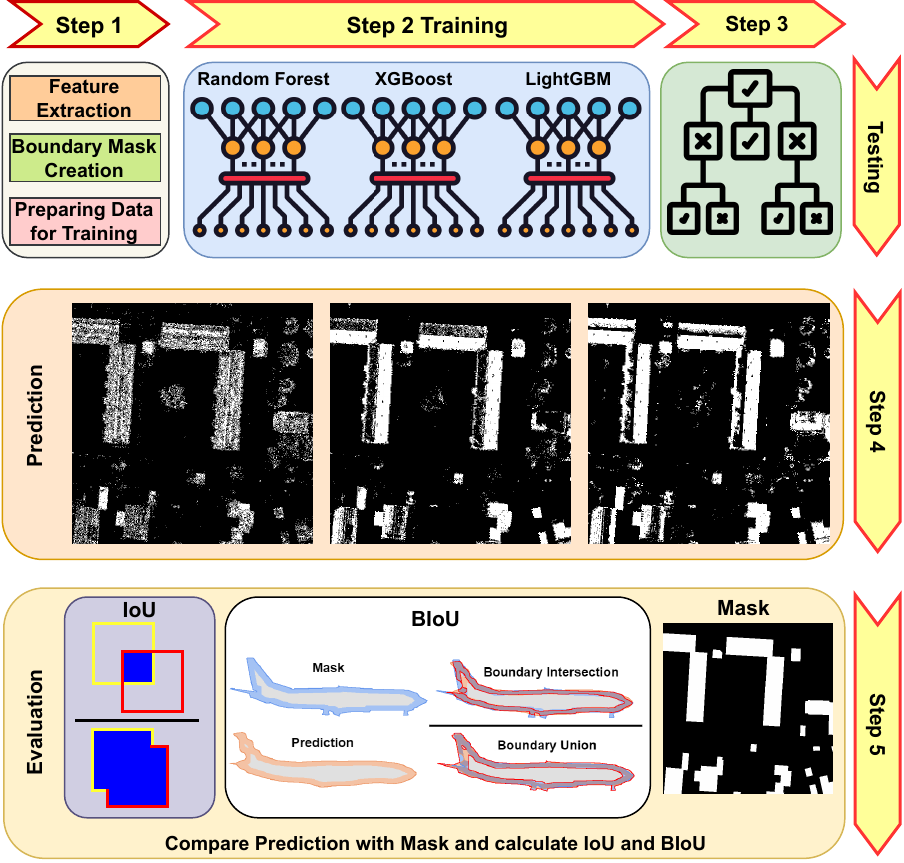}
\caption{Segmentation: Preprocessing, Training, Testing, Prediction, and Performance Evaluation.} \label{segmentation}
\end{figure}

To ensure how LiDAR data affect the results of pixel-wise segmentation, initially, only aerial images are used, and later LiDAR data along with aerial images are used in experiments. In the first task, only four features (blue, green, red, and gray) are used in training, validation, and testing. In the second task, LiDAR data is also used to exploit the presence of LiDAR data for segmentation. In Figure \ref{segmentation} step 1 presents feature extraction and preprocessing of the data for segmentation. In this step, features are extracted from the images and the image-based dataset is prepared to be trained on traditional machine learning algorithms. 

In the second step, three classifier: RF, XGBoost and Light GBM are trained on the data. In the case of the deep learning model, a complete image is used as an input to the model for training and provides segment maps as output. This segment map consists of $n$ number of channels, one for each label class. Contrary to the deep learning decision, the classifiers used in this work fed up with one-pixel information at each time for model training. Trained models are later stored for validation testing purposes.

In the third step, trained models are tested on testing data of the dataset. Feature of each image are extracted, preprocessed, and later pixel-wise tested on the stored models. Models predicted each pixel value either as a building or foreground. In fourth step, the output of testing are reasembled as a predicted mask. In Figure \ref{segmentation}, prediction component represent predicted mask for RF, XGBoost and LightGBM.

In the fourth step, Evaluation performs with the help of IoU and BIoU by comparing the ground truth mask with the predicted mask. In IoU intersection of both mask is divided by the union of both masks, while in BIoU, the boundary intersection is divided by boundary union. IoU validate raw accuracy of the model, while BIoU validates how perfectly the contour of the building is segmented by the model. BIoU is more sensitive toward the shape of the building, which is more challenging as compared to IoU.

%% file: chapters/chapter4.tex


\vspace{0.5em}
Table \ref{ParameterSetup} show the parameter setup for the experiment done in this work. All features mentioned in the table are tested on the given dataset. Features are listed in the table according to models used for interpretability. The best performance was achieved using Blue, Green, Red, and Gray as shown with bold text. The remaining experiments are performed using only these best features. RF classifier tested on the mentioned estimator values and $10$ show best score for the dataset. All experiments shown in Table \ref{Randomforest} are performed with best estimator. Hyperparameter for XGBoost and LightGBM obtain from tuning on the given dataset using grid search CV. Further more learning rate form LightGBM test from 0.05 to 0.0009 but the impact on the dataset is very minimal as compared to computation increase for training.

\begin{table}
\centering
\caption{Parameter setup for experiments.}
\label{ParameterSetup}
\setlength{\tabcolsep}{2pt}
\begin{tabular}{|c|c|} 
\hline
Classifier & Parameter \\ 
\hline
Features   & \begin{tabular}[c]{@{}c@{}}\textbf{Blue}, \textbf{Green}, \textbf{Red}, \textbf{Gray}, Histogram Equalization,\\~Morphological, Clahe Histogram Equalization,\\~Gabor Filter, Canny Edge Detector\end{tabular}  \\ 
\hline
RF         & n\_estimators (3,4,5,6,8,\textbf{10},12) \\ 
\hline
XGBoost    & \begin{tabular}[c]{@{}c@{}}colsample\_bytree:0.9, gamma:8.3, max\_depth:8,\\~min\_child\_weight:5, reg\_alpha:177, reg\_lambda:0.04\end{tabular}\\ 
\hline
LightGBM   & \begin{tabular}[c]{@{}c@{}}learning\_rate:(0.05-0.0009), boosting\_type:gbdt,\\~objective:binary,~~metric:[auc, binary\_logloss], \\~num\_leaves:100, max\_depth:10\end{tabular}                                \\
\hline
\end{tabular}
\end{table}

\begin{table*}
\centering
\caption{Evaluation Analysis}\label{Randomforest}
\begin{tabular}{|c|c|c|c|c|c|c|c|}
\hline
 & & & & \multicolumn{2}{|c|}{\textbf{Task 1}} & \multicolumn{2}{|c|}{\textbf{Task 2}}  \\\hline
\textbf{Classifier} & \textbf{Features} & \textbf{Images} & \textbf{BMask} & \textbf{IoU} & \textbf{BIoU} & \textbf{IoU} & \textbf{BIoU} \\\hline
\multirow{6}{*}{RF} & BGRGray(L) & 10 & No & 0.8711	& 0.273 & 0.8812 & 0.3009 \\
 & BGRGray(L) & 100 & No & 0.9037	& 0.3235 & 0.9190 & 0.3389 \\
 & BGRGray(L) & 1000 & No & 0.9237	& 0.3300 & 0.9317 & 0.3445 \\
 & BGRGray(L) & 2000 & No & 0.9235	& 0.3350 & 0.9318 & 0.3444 \\
 & BGRGray(L) & 7000 & No & 0.9234	& 0.3348 & 0.9321 & 0.3436 \\
 & BGRGray(L) & 2000 & Yes & 0.8562	& 0.5460 & 0.8763 & 0.5562 \\\hline

\multirow{6}{*}{XGBoost} & BGRGray(L) & 100 & No & 0.9011	& 0.2876 & 0.8912 & 0.3010 \\
 & BGRGray(L) & 500 & No & 0.8837	& 0.3335 & 0.8854 & 0.3589 \\
 & BGRGray(L) & 1000 & No & 0.8843	& 0.3501 & 0.8817 & 0.3545 \\
 & BGRGray(L) & 7000 & No & 0.8823	& 0.3513 & 0.8821 & 0.3535 \\
 & BGR(L) & 1000 & No & 0.8833	& 0.3654 & 0.8845 & 0.3876 \\
& BGR(L) & 1000 & Yes & 0.8802	& 0.5401 & 0.8799 & 0.5672 \\\hline

\multirow{6}{*}{LightGBM} & BGRGray (L) & 10 & No & 0.8635	& 0.4050 & 0.8782 & 0.424 \\
& BGRGray(L) & 10 & Yes & 0.8933	& 0.443 & 0.8945 & 0.464 \\
 & BGRGray(L) & 100 & Yes & 0.8831	& 0.5221 & 0.8851 & 0.5331 \\
 & BGRGray(L) & 1000 & Yes & 0.8763	& 0.5831 & 0.8783 & 0.5998 \\\hline

\end{tabular}
\end{table*}

RF is a slow classifier as compared to XGBoost and LightGBM; it requires more data for training. As the number of images increases, performance also improves, but after training on 2000 images, performance remains the same. Table \ref{Randomforest} shows performance on 7000 is almost the same as 2000 images.  In feature column (L), denote LiDAR data for task 2. Column BMask denotes boundary mask and RF also tested with boundary masks. IoU dropped 5\% while BIoU increased 22\% in task 1 and almost the same pattern was shown in task 2. Comparison analysis of tasks 1 and 2 shows that including LiDAR data improves both IoU and BIoU scores.

As compared to RF, XGBoost required fewer data for training. The best performance was achieved with 1000 images for training and was not improved further using all images. Table \ref{Randomforest} shows better results for XGBoost when the gray channel is not used as a feature due to it less influence on the output label. Same to RF, the inclusion of boundary masks improves the performance score, and LiDAR data with aerial images in each experiment perform better as compared to only aerial images. The performance score of XGBoost is better on both metrics as compared to RF.

LightGBM can be quickly trained on fewer data as compared to RF and XGBoost. Table \ref{Randomforest}, shows results for XGBoost on only 10 images with and without boundary mask, which is relatively better as compared to RF and XGBoost. As the number of images increases, BIoU also increases. In LightGBM, feature ranking is not possible, due to which the same four features for task one and five features for task 2 are used. The inclusion of boundary mask improves BIoU, same in RF and XGBoost. Same as other models, LiDAR data improve both IoU and BIoU.

\begin{table*}
\centering
\caption{Comparison of RF, XBGoost and LightGBM}\label{Model Comparison}
\setlength{\tabcolsep}{8pt}
\begin{tabular}{|c|c|c|c|c|c|c|c|}
\hline
 & \multicolumn{3}{|c|}{\textbf{Task 1}} & \multicolumn{3}{|c|}{\textbf{Task 2}} &  \\\hline
 \textbf{Classifier} & \textbf{IoU} & \textbf{BIoU} & \textbf{Total} & \textbf{IoU} & \textbf{BIoU} & \textbf{Total}& \textbf{Score} \\\hline
 
 RF & 0.8562	& 0.5460 & 0.7011 & 0.8763 & 0.5562  & 0.7162 & 0.7086 \\\hline
 XGBoost & \textbf{0.8802}	& 0.5401 & 0.7101 & \textbf{0.8799} & 0.5672  & 0.7235 & 0.7123 \\\hline
 LightGBM & 0.8763	& \textbf{0.5801} & \textbf{0.7282} & 0.8783 & \textbf{0.5998}  & \textbf{0.7390} & \textbf{0.7336} \\\hline
\end{tabular}
\end{table*}

Table \ref{Model Comparison} compares models used in this work. Total is the average of IoU and BIoU in both tasks. The score is the average of both totals in the table. The IoU score of XGBoost is better in task 1 and task 2, while BIoU score of LightGBM is significantly better as compared to others. LightGBM performs better in average scores for both tasks and also overall scores.

\begin{table}
\centering
\caption{Comparison with top 3 competitors from MAPAI Competition}\label{Comparison}
\setlength{\tabcolsep}{5pt}
\begin{tabular}{|c|c|c|c|c|}
\hline
 & \multicolumn{2}{|c|}{\textbf{Task 1}} & \multicolumn{2}{|c|}{\textbf{Task 2}} \\\hline
 \textbf{Classifier} & \textbf{IoU} & \textbf{BIoU}  & \textbf{IoU} & \textbf{BIoU}  \\\hline
 
 FUNDATOR \cite{hodne2022team} & 0.7794 & 0.6115	& 0.8775 & \textbf{0.7857 } \\\hline
 HVL-ML \cite{kaliyugarasan2022lab} & 0.7879 & \textbf{0.6245} & 0.8711	& 0.7504 \\\hline
DEEPCROP \cite{li2022buildseg} & 0.7902 & 0.6185 & 0.8506 & 0.7461\\\hline
 Proposed & \textbf{0.8763}	& 0.5831 & \textbf{0.8783} & 0.5998\\\hline
\end{tabular}
\end{table}

Table \ref{Comparison} compares proposed work with the top 3 competitors from NORA MapAI competition. The proposed work is based on pixel-wise segmentation by neglecting the context of the pixel with its neighbors, which results in less score for BIoU metrics in both tasks. While proposed work claims a better IoU score as it focuses more on the values of each pixel rather than the context of the pixel relative to neighbors. The segmentation method of all top three competitors is based on a deep learning model which accounts the contextual segmentation rather than pixel-wise and hence shows a better BIoU score as compared to the proposed one.

\begin{figure*}
\centering
\includegraphics[width=\textwidth]{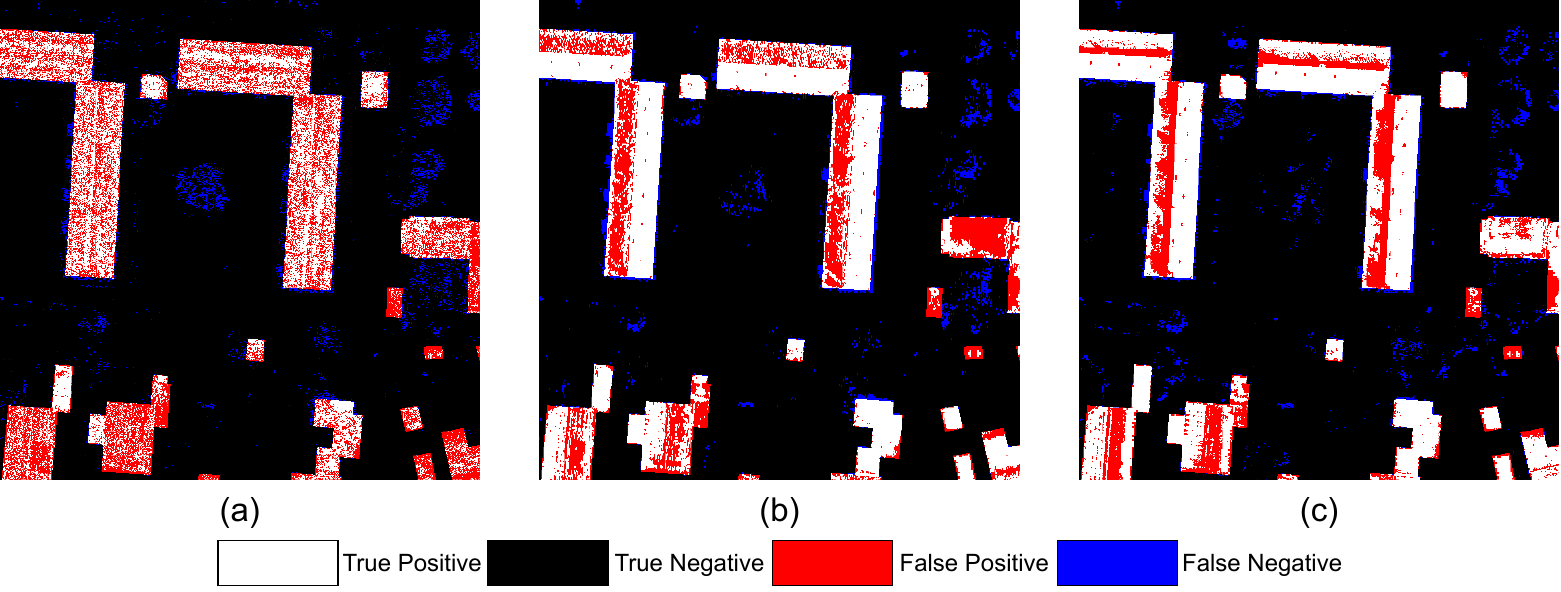}
\caption{Classifiers Mask (a) RF (b) XGBoost and (c) LightGBM} \label{classifiermask}
\end{figure*}

Figure \ref{classifiermask} shows generated masks for RF, XGBoost, and LightGBM. RF mask has more red and blue pixels as compared to others, and building roofs could be enhanced with a median filter which can improve IoU, but it will disturb true positive pixels at the edges, which will result in less BIoU score. XGBoost performs better as compared to Random; both false negative and false positive are decreased, and roofs illuminated by light are predicted better. LigthGBM further improves the prediction by reducing false positives as compared other two models.

\begin{table}
\centering
\caption{Interpretability of RF and XGBoost}\label{interpretabiliity}
\setlength{\tabcolsep}{5pt}
\begin{tabular}{|c|c|c|c|c|}
\hline
 & \multicolumn{2}{|c|}{\textbf{RF}} & \multicolumn{2}{|c|}{\textbf{XGBoost}} \\\hline
 \multirow{4}{*}{\textbf{Task 1}} & Blue & 0.368099
  & Blue & 0.38609\\
  & Red & 0.299738	& Red & 0.30973 \\
  & Green & 0.190407 & Green	& 0.200707 \\
 & Gray & 0.141755 & Gray & 0.102655\\\hline
\multirow{5}{*}{\textbf{Task 2}} & LiDAR & 0.261568
  & LiDAR & 0.233052\\
  & Blue & 0.209854	& Blue & 0.213057 \\
  & Red & 0.197456 & Green	& 0.207829 \\
 & Green & 0.160032 & Red & 0.195888\\
 & Gray & 0.170462 & Gray & 0.150174\\\hline
\end{tabular}
\end{table}

Table \ref{interpretabiliity} show the interpretability of RF and XGBoost model for both task, while LightGBM exhibits the interpretable nature same as the deep learning model. This table shows, how much significant (a high number show more significance and vice versa) these features are to predict precisely the output. In task 1, blue and red are the most significant in both RF and XGBoost. Gray show more significance in RF model as compared to the XGBoost model. The main aim of using shallow models in this work is to determine the significance of LiDAR data compared to aerial images. In task 2, LiDAR data is the most significant feature for both RF and XGBoost models leaving behind the aerial images features that are blue, red, green and gray channels of the image.

%% file: chapters/chapter5.tex
\vspace{0.5em}
In NORA competition, different deep learning models are tested on the MapAI dataset, though the problem belongs to binary segmentation. Due to the problem's simplicity and interpretability, this work targeted boosting models as shallow learning models to perform binary segmentation on the given dataset. It was expected to achieve the same, if not good, results as compared to the deep learning model, but due to the nature of different evaluation metrics, the proposed work outperformed deep learning in IoU. In contrast, deep learning models have better scores in BIoU.

This work focuses on pixel-wise segmentation rather than segmentation using a deep learning algorithm.
Like the NORA MapAI competition, this work also uses IoU and BIoU to evaluate the models. proposed work results in better IoU than competitors in Nora competition as it focuses more on pixel information individually for segmentation. BIOU is worst compared to the competition because RF, XGBoost, and LightGBM are not considering the shapes of the building due to their nature. Table \ref{interpretabiliity} shows that LiDAR data is more significant as compared to other features. However, to be interpretable, they pay a price in terms of prediction accuracy as compared to deep learning models. While the deep learning model extracts patterns from the image using convolution, each pattern consists of a group of pixels related to each other. Hence deep learning is better is segmenting objects and shape in the image, which result in better BIoU as compared to the proposed work.